\documentclass[conference]{IEEEtran}
\IEEEoverridecommandlockouts

\usepackage[utf8]{inputenc}
\usepackage[english]{babel}
\usepackage{csquotes,xpatch}
\usepackage[backend=biber,style=ieee]{biblatex}
\usepackage{amsmath,amssymb,amsfonts}
\usepackage{algorithmic}
\usepackage{graphicx}
\usepackage{textcomp}
\usepackage{xcolor}
\usepackage{float}
\usepackage[ruled,vlined]{algorithm2e}
\usepackage{cleveref}

\providetoggle{blx@lang@captions@english}

\def\BibTeX{{\rm B\kern-.05em{\sc i\kern-.025em b}\kern-.08em
    T\kern-.1667em\lower.7ex\hbox{E}\kern-.125emX}}
    
\begin{document}

\title{Indonesian ID Card Extractor Using Optical Character Recognition and Natural Language Post-Processing\\

\thanks{IDentify applicable funding agency here. If none, delete this.}
}

\author{\IEEEauthorblockN{1\textsuperscript{st} Firhan Maulana Rusli}
\IEEEauthorblockA{\textit{School of Computing} \\
\textit{Telkom University}\\
Bandung City, West Java, Indonesia \\
firhanmaulanarusli@gmail.com}
\and
\IEEEauthorblockN{2\textsuperscript{rd} Kevin Akbar Adhiguna}
\IEEEauthorblockA{\textit{Faculty of Mathematics and Natural Science} \\
\textit{Padjadjaran University}\\
Sumedang district, West Java, Indonesia \\
kevin17016@mail.unpad.ac.id}
\and
\IEEEauthorblockN{3\textsuperscript{nd} Hendy Irawan}
\IEEEauthorblockA{\textit{School of Computing} \\
\textit{Telkom University}\\
Bandung City, West Java, Indonesia \\
hendy@lovia.life}
\and

}

\maketitle

\begin{abstract}
The development of Information Technology has been increasingly changing the means of information exchange leading to the need of digitizing print documents. In the present era, there is a lot of fraud that often occur. To avoid account fraud there was verification using ID card extraction using OCR and NLP. Optical Character Recognition (OCR) is technology that used to generate text from image.
With OCR we can extract Indonesian ID card or \emph{kartu tanda penduduk} (KTP) into text too. This is using to make easier service operator to do data entry.
To improve the accuracy we made text correction using Natural language Processing (NLP) method to fixing the text. With 50 Indonesian ID card image we got 0.78 F-score, and we need 4510 milliseconds to extract per ID card.

\end{abstract}


\section{Introduction}
The development of Information Technology has been increasingly changing the means of information exchange leading to the need of digitizing print documents \autocite{1}. In the present era, there is a lot of fraud that often occur, there are so many types of frauds in this era, for example is account fraud. ID card in Indonesia is KTP. The Kartu Tanda Penduduk (literally: Resident identity Card), commonly KTP is an Indonesian compulsory identity card \autocite{2}. To avoid account fraud there was verification using ID card or KTP, then service operator extract the data in ID card. All identity in ID card, confidence for every field and face photo encoder field. But usually service operator extract the ID card in conventional way generally referred as data entry.

Service operator or someone who verifying the ID card manually is normally run by humans. Human need to take rest, human can't work 24 hours without sleep and sometimes humans made mistakes when doing the data entry because there were too much ID card that should be entry or to write down the ID card content.

Because of human sometimes made mistake, then we made automation to do data entry ID card using Optical Character Recognition (OCR) and post processing using Natural Language Processing (NLP). This technique have been developed to transform ID card into digital documents. However, various image, with good and bad quality, become challenge for OCR engine \cite{3}. 

We use OCR to extract ID card image into digital documents and because sometimes OCR got and error or bad result \cite{3}, we use Natural Language Processing (NLP) for fixing text.

This research is in the form of answering the following questions about the development of the Indonesian ID card extraction system:

\begin{enumerate}   
    \item How OCR and NLP works in ID card extractor?
    \item How fast is the processing per ID card?
    \item The F-score of ID card extractor using OCR and NLP?
\end{enumerate}

Our analysis should be beneficial for researchers and practitioners helping them better understand strengths as well as weakness of this approach.

The remainder of this paper is organized as follows. We introduce literature review in Sec. 2. Then, Sec. 3 dataset and methodology. In Section 4, result and discussion. After that, the summary of our major findings is shown in Section 5.

\section{Literature Review}

\subsection{ID Card}
Indonesian ID Card can be used to recognize
citizen of Indonesia Identity in several requirements like for
sales and purchasing recording, admission and other
transaction processing systems (TPS) \cite{4}. Indonesian ID card or KTP had 37 fields. 

\begin{figure}[H]
    \centering
    \includegraphics[scale=0.15]{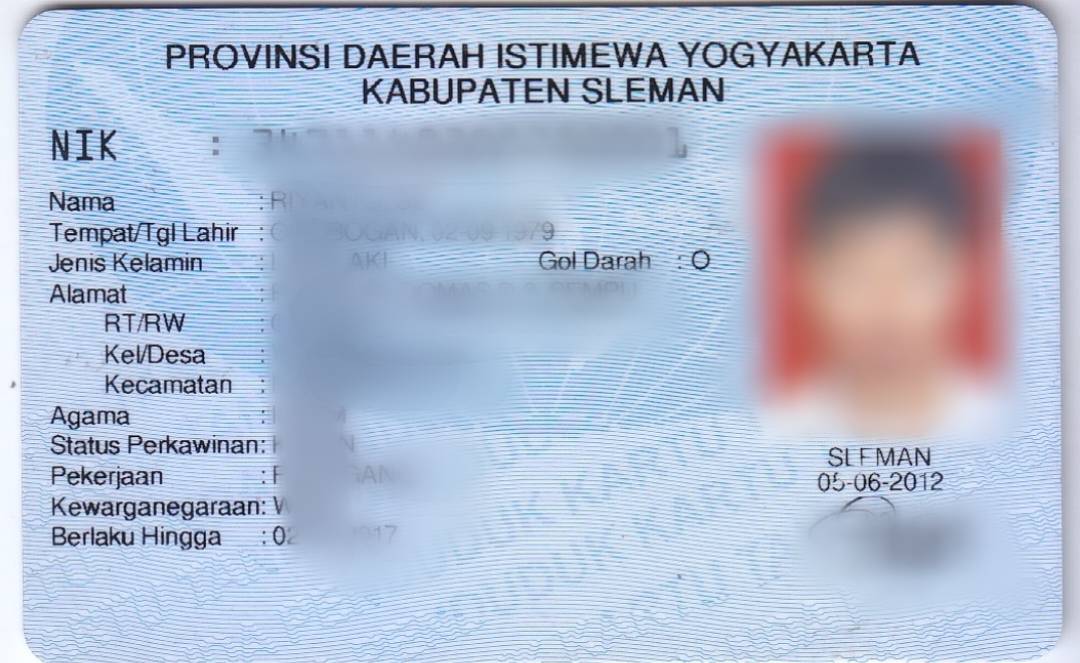}
    \caption{Fields in KTP or ID card}
    \label{fig1}
\end{figure}

\cref{fig1} shows that KTP or Indonesian ID card had 37 fields. There are many things we can do with ID card such as borrowing money, create various kinds of documents like driver's license, passport and many more. That's why we make Id card extractor to speed up the process, because there were so many instances that still almost on their input process of ID card was done by using conventional way. 

\subsection{Optical Character Recognition}
Optical Character Recognition (OCR) is a process of converting a machine-printed or handwritten text image into a digital computer format that can be editable, OCR technology is considered as a challenging research area in the field of pattern recognition and artificial intelligence \cite{5}. This technology is often called text detection. OCR is part of an automatic identification technique. Sometimes the traditional way to input data via keyboard is not the most efficient way. There are several library in OCR, there were PyTesseract, TesseOCR, PyOCR and PyTesseract is the highest score for OCR library based on \cref{table_ocr_result} \cite{6}.

\begin{table}[htbp]
\caption{OCR Library Result}
\begin{center}
\begin{tabular}{||c c c||} 
 \hline
 Library & Speed & Total Area   \\ [0.5ex]
 \hline\hline
 PyOCR & 24.7S & 0.36  \\
 PyTessseract & 25.16S & 0.46 \\ 
 TesseOCR & 22.53S & 0.30 \\ [1ex] 
 \hline
\end{tabular}
\label{table_ocr_result}
\end{center}
\end{table}

PyTesseract had another attribute called confidence, confidence is using for how sure the word have been extracted. We count the confidence for every field, these results are used for service operator later.

\begin{figure}[H]
    \centering
    \includegraphics[scale=0.45]{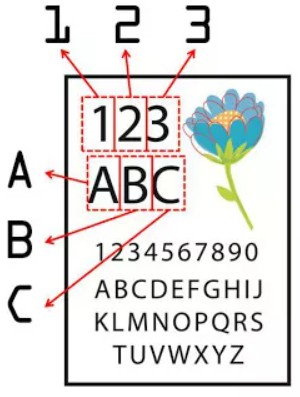}
    \caption{Matching Every Character \cite{marcom1-digitalsense_fitur_2015}}
    \label{fig2}
\end{figure}

\cref{fig2} shows that OCR works by matching every letter to the close's character \autocite{marcom1-digitalsense_fitur_2015}. For increasing the accuracy, we use preprocessing, first change the image into grayscale and after that change the grayscale into binary. 

\begin{table}[htbp]
\caption{F-score Evaluation \cite{4}}
\begin{center}
\begin{tabular}{||c c c||} 
 \hline
 No & Model & F-Score  \\ [0.5ex] 
 \hline\hline
 1 & Convolutional Neural Network & 0.84  \\ 
 2 & Support Vector Machine & 0.63  \\ [1ex] 
 \hline
\end{tabular}
\label{table2}
\end{center}
\end{table}

\cref{table2} shows that Convolutional Neural Network is better than Support Vector Machine for extracting Indonesian ID card \cite{4}. We will compare CNN with this research. We are using OCR and post-processing using Natural Language Processing for ID card extraction.

\section{Methodology}

\begin{figure}[H]
    \centering
    \includegraphics[scale=0.54]{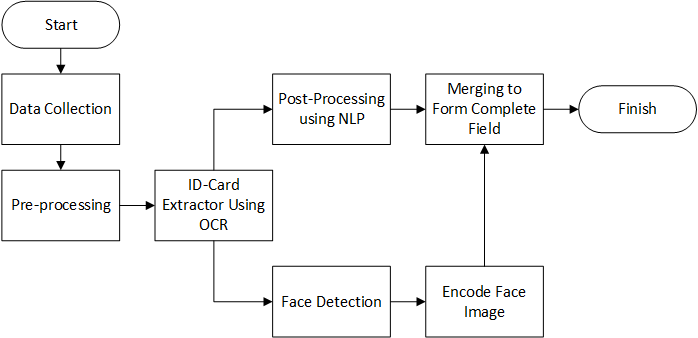}
    \caption{System Flowchart}
    \label{fig}
\end{figure}

In this research we developed ID card extraction with OCR and post-processing using NLP. There were 37 fields in Indonesian ID card, there are confidence fields, face encoder field, all identity in ID card, confidence for every field and face photo encoder field. For the analysis, we utilize 50 Indonesian ID card image. Those datasets were collected from public using Google Form submission. Then those datasets we preprocess, after that we extract it using OCR, then process again the result using NLP and using face detection on ID card to extract the face on ID card, and the last step merging all those data to form.

\subsection{Data Preprocessing}
\begin{figure}[H]
    \centering
    \includegraphics[scale=0.25]{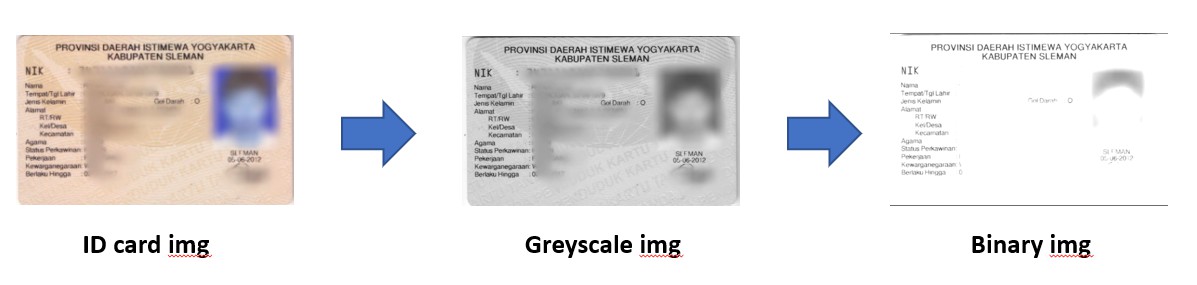}
    \caption{Processing IMG from RGB too Grayscale to Binary}
    \label{fig3}
\end{figure}

\cref{fig3} shows that dataset be processed first using Grayscale technique to convert RGB layer into Grayscale, then threshold is performed to convert grayscale image \(R(x,y)\) into binary by selecting apropriate threshold, Tr means threshold , it set by 127, F(x,y) means the binary image \cref{eq1}.

\begin{equation}
F(x,y) 
        = \left\{ 
        \begin{array}{l} 
            1, if \quad R(x,y) > Tr\\
            0, Otherwise\\
        \end{array}
        \right.
\label{eq1}
\end{equation}

\subsection{Extracting ID Card}
\begin{figure}[H]
    \centering
    \includegraphics[scale=0.3]{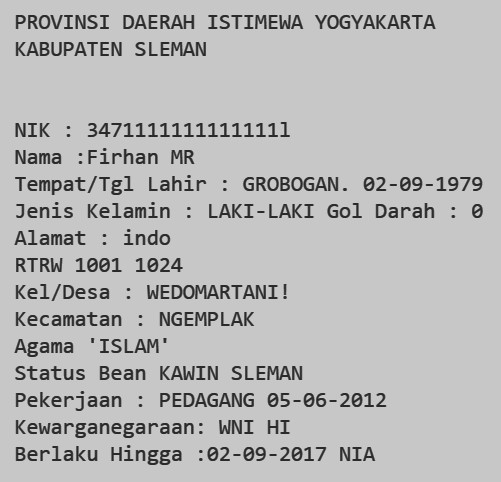}
    \caption{ID card Extraction Result}
    \label{fig5}
\end{figure}

In this system for the OCR process a module on
python programming. The module is called Python Tesseract which is an OCR module that is integrated with python in the form of a library, after all preprocessing steps are carried out to obtain an image in binary form. Then the image is processed using the Pytesseract library (Python Tesseract) so that text data is obtained from the image captured by the camera. Because we extracted Indonesian ID card then we add "lang = 'ind'" in hyperparameter tesseract. \cref{fig5} shows the result of ID card extractor.

\subsection{Post-Processing}
After we got the result from extracting image \cref{fig5}, there would be wrong spelling like field "Gol Darah : 0" the content of attribute "Gol Darah" is not "0" instead "O", and field "Kel/Des : WEDOMARTANI!" it instead be "WEDOMARTANI" without "!" symbol, because there are wrong spelling in result of ID Card extractor then we are using NLP to fixing it. First we are using Punctual Remover on the result, but we don't remove ":.,-" symbol, because that symbol is part of content in ID card. After that we are using RegEx. Regex is Regular Expression, Regex is using for finding sequence of characters that define a search pattern \autocite{7}. Then we are using word to number converter, we are using this for "NIK" field, because there are misspelling character because scratch ID card or bad quality image, so we make function to take care of it, word to number converter list:

\begin{itemize}
    \item "L" : "1",
    \item 'l' : "1",
    \item 'O' : "0",
    \item 'o' : "0",
    \item '?' : "7",
    \item 'A' : "4",
    \item 'Z' : "2",
    \item 'z' : "2",
    \item 'S' : "5",
    \item 's' : "5",
    \item "b" : "6",
    \item "B" : "8",
    \item "G" : "6"
\end{itemize}

\begin{figure}[H]
    \centering
    \includegraphics[scale=0.5]{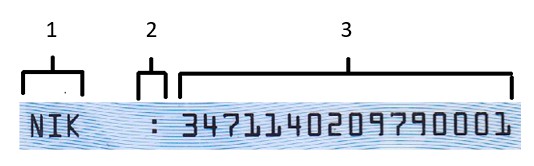}
    \caption{3 Part of Every Field}
    \label{fig9}
\end{figure}

After we are splitting the sentence by ":" to take every content of field, because we just need the content not the attribute. for example "Name : Firhan Maulana", we just need "Firhan Maulana" not "Name :". every part of field is divided by three part, shows on \cref{fig9}.

\begin{figure}[H]
    \centering
    \includegraphics[scale=0.7]{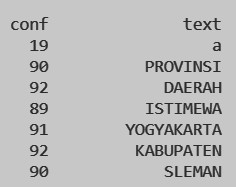}
    \caption{Example Confidence Word by Word}
    \label{fig7}
\end{figure}

Part 1 is attribute, part 2 is colon and part 3 is the content. Then we just use RegEx to find attribute on every field, and then split it by colon, and then take the content which is the last part. But there was special case for some field. For "Tempat/tanggal lahir" we are using this algorithm. re.search('([0-9]{2}\-[0-9]{2}\-[0-9]{4})') this algorithm is using to search pattern date on "Tempat/Tanggal Lahir" field. For "Jenis Kelamin" field, we are using library Regex.search, because "Jenis kelamin" or Gender just have two types, it were "LAKI-LAKI" and "PEREMPUAN" then using this algorithm, re.search("(LAKI-LAKI|LAKI|LELAKI|PEREMPUAN)"), re.search is using too for other field that containing limited type of content, there were "Agama", "Golongan Darah", "Kewarganegaraan" and "Status", because service operator need proof that this ID card extraction result is truly got the same content equal to the real ID card. Then we added the confidence for every field. Confidence is how true the image extracting to text presented by scale 0 to 100. Confidence one of the modules in Pytesseract library. Actually confidence counting percent word-by-word not per-line shows by \cref{fig7}. 

Because we want to count the confidence by per-line then we gather word by word to sentence then calculate the mean of sentence then it becomes counting confidence per-line, or we can call it per-field. We count this confidence by break down every field in result, then take the confidence word, and calculate the mean of the field. For Example for “Provinsi” field we got confidence 91, this result by calculating mean of confidence show on \cref{table3}.

\begin{table}[htbp]
\caption{Example Confidence per-word}
\begin{center}
\begin{tabular}{||c c c c||} 
 \hline
 Province & DAERAH & ISTIMEWA & YOGYAKARTA  \\ [0.5ex] 
 \hline\hline
 Conf & 92 & 89 & 91  \\  [1ex] 
 \hline
\end{tabular}
\label{table3}
\end{center}
\end{table}

\subsection{Face Detection}
\begin{figure}[H]
    \centering
    \includegraphics[scale=0.45]{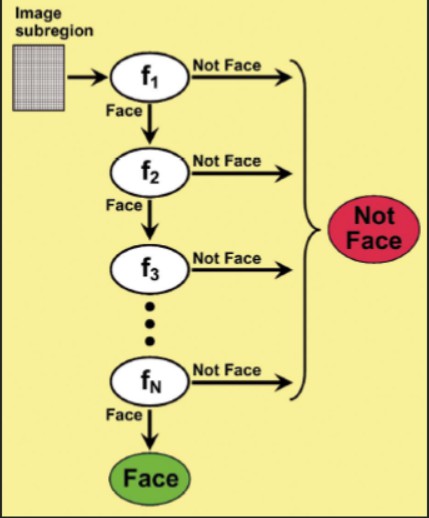}
    \caption{Chain of Filters \cite{noauthor_face_2012}}
    \label{fig8}
\end{figure}

Face detection is using for detect face on ID card, this because some of the service operator need to take the face photo on the ID card. So we take the face photo using Haarcascade Frontalface Default, it was open source from OpenCv \autocite{goel_hybrid_2012}. This feature uses the Haar waveform. The Haar waveform is a square wave. In 2 dimensions, a square wave is a pair of adjacent squares, 1 light, and 1 dark. Haar is determined by subtracting the mean dark area pixels from the average light area pixels \autocite{noauthor_face_2012}. If the difference is above the threshold, the feature is said to be present. To determine whether there is a Haar feature at each image location, this technique is called Integral Image. Generally, integrals add up in small units. In this case, this small unit is called the pixel value. The integral value for each pixel is the sum of all the pixels above and to the left. From top left to bottom right, images can be integrated as per pixel mathematical operations. Filters at each level are trained to classify images that have been filtered previously. During use, if one of the filters fails, the image region in the image is classified as “Not Face”. When the filter succeeds in passing the image region, the image region is then included in the next filter \autocite{noauthor_face_2012}. Image regions that have gone through all filters will be considered as “Faces”, shows on \cref{fig8}.

After got the face photo, we encode the image so it becomes string then we can add to field "facephoto". We encode the face photo using Base64 encoding.

\section{Result and Discussion}

Example for final result would be like this:

\begin{enumerate}
    \item "kind": "C", 
    \item "identifier": "3471111111111111", 
    \item "name": "Firhan Maulana", 
    \item"birthPlace": "GROBOGAN. ", 
     \item"birthDate": "02-09-1979", 
     \item"gender": "M", 
     \item"bloodType": "O", 
     \item"address": "PRM PURI DOMAS RT : 001 RW : 024 KELURAHAN/DESA : WEDOMARTANI KECAMATAN : NGEMPLAK",
     \item"religion": "ISLAM", 
     \item"marriageStatus": "M", 
     \item"occupation": "PEDAGANG", 
     \item"nationalityCode": "IND", 
     \item"expiryDate": "SEUMUR HIDUP", 
     \item"facePhoto":"2KfZhNiz2YTYp9m...", 
     \item"cardImage": "Yg2KfZhNmk/Ysf...", 
     \item"issuerCountryCode": "IND", 
     \item"issuedProvince": "JAKARTA", 
     \item"issuedCity": "KABUPATEN SLEMAN", 
     \item"issuedDate": "05-06-2012", 
     \item"faceTop": 786, 
     \item"faceLeft": 212, 
     \item"faceWidth": 163, 
     \item"faceHeight": 163, 
     \item"extractedAt": "30-11-2020", 
     \item"identifierconf": 0, 
     \item"nameconf": 91, 
     \item"birthPlaceconf": 54, 
     \item"birthDateconf": 95, 
     \item"genderconf": 71, 
     \item"bloodTypeconf": 76, 
     \item"addressconf": 68.0, 
     \item"religionconf": 18, 
     \item"marriageStatusconf": 95, 
     \item"occupationconf": 95, 
     \item"issuedProvinceconf": 95, 
     \item"issuedCityconf": 95, 
     \item"issuedDateconf": 83
\end{enumerate}

ID Card Extractor give output as Json file with 37 fields, it becomes 37 fields because extract the face photo in the ID card too and give the confidence for every field. Extracting ID card with OCR have common errors or false spelling like colon after word, slash symbol and usually errors come from symbol in ID card. Quality of image will reduce OCR accuracy this are because OCR read pattern of pixels and decide on the closest match of characters. Bad quality photo or old ID card that sometimes had scratch can have an effect on the quality of the result.
This confidence is using for service operator later, if confidence more than 85 for one field, then admin does not have to double check the result. For example "NAMA : Firhan Maulana" got confidence 91, then service operator does not have to double check, if it lower than 85 then service operator have to double check the result, because when the confidence lower than 85 there were misspelling on the field.

ID card extractor using OCR and post processing using NLP have 0.78 F-score total, it was divided by two type of photo, 25 for ID card using camera with 0.67 F-score, and 25 photo using scanner with 0.89 F-score. For measurement using F-score, the complete result can be seen at \cref{table4}, CNN result still better than just using OCR and Post processing using NLP, and we need 4510 millisecond on average to extract per ID card.

\begin{table}[htbp]
\caption{Final Result}
\begin{center}
\begin{tabular}{||c c c||} 
 \hline
 No & Model & F-Score  \\ [0.5ex] 
 \hline\hline
 1 & Using OCR and Post processing using NLP & 0.78 \\
 2 & Convolutional Neural Network & 0.84  \\ 
 3 & Support Vector Machine & 0.63  \\ [1ex] 
 \hline
\end{tabular}
\label{table4}
\end{center}
\end{table}

\section{Conlusion}
We created ID card extractor using OCR and post-processing using NLP with 50 dataset ID card that divided by two type, 25 for scanning image, and 25 for camera image. F-score that we have obtained was 0.78 for all image, is divided by 0.89 F-score for scanning image and 0.67 F-score for camera image. From previous research using CNN model 0.84 F-score, and we need 4510 milliseconds to extract per ID card.

For the future experiment, we plan to use deep leaning Recurrent Neural Networks (RNN),Residual Neural Networks (ResNet),  convolutional Neural Network (CNN). Then using NLP for the post-processing.

\section*{Acknowledgment}

First we would like to thank Lovia (https://about.lovia.life), who give us facilities to finish this research paper.
Second we would like to thank to our parents, who supported us with love and understanding. Without them, we could never have reached this current level of success. 


\printbibliography[]

\end{document}